\documentclass[sn-apa,iicol]{sn-jnl}% APA Reference Style
\usepackage{blindtext}
\usepackage{subcaption}

\renewcommand{\vec}[1]{\boldsymbol{#1}}
\renewcommand{\vec}[1]{\boldsymbol{#1}}
\renewcommand{\set}[1]{\mathcal{#1}}

\renewcommand{\x}{\vec{x}}
\renewcommand{\b}{\vec{b}}
\renewcommand{\p}{\vec{p}}
\renewcommand{\y}{y}

\newcommand{\detectorParameters}{\vec{\omega}}
\newcommand{\detector}{f^{\detectorParameters}}
\newcommand{\simplex}[1]{\Delta_{#1}}
\newcommand{\background}{\otimes}

\jyear{2021}%

\theoremstyle{thmstyleone}%
\theoremstyle{thmstyletwo}%

\theoremstyle{thmstylethree}%

\begin{document}

\title[Uncertainty in Object Detectors]{A Review of Uncertainty Calibration in Pretrained Object Detectors}


\author*[1]{\fnm{Denis} \sur{Huseljic}}\email{dhuseljic@uni-kassel.de}

\author[1]{\fnm{Marek} \sur{Herde}}\email{marek.herde@uni-kassel.de}

\author[1]{\fnm{Mehmet} \sur{Muejde}}\email{mehmet.muejde@uni-kassel.de}

\author[1]{\fnm{Bernhard} \sur{Sick}}\email{bsick@uni-kassel.de}
%\equalcont{these authors contributed equally to this work.}

\affil*[1]{\orgdiv{Intelligent Embedded Systems}, \orgname{University of Kassel}, \orgaddress{\street{Wilhelmshöher Allee 73}, \city{Kassel}, \postcode{34121}, \state{Hessen}, \country{Germany}}}

\abstract{
In the field of deep learning based computer vision, the development of deep object detection has led to unique paradigms (e.g., two-stage or set-based) and architectures (e.g., \textsc{Faster-RCNN} or \textsc{DETR}) which enable outstanding performance on challenging benchmark datasets. Despite this, the trained object detectors typically do not reliably assess uncertainty regarding their own knowledge, and the quality of their probabilistic predictions is usually poor. As these are often used to make subsequent decisions, such inaccurate probabilistic predictions must be avoided. In this work, we investigate the uncertainty calibration properties of different pretrained object detection architectures in a multi-class setting. We propose a framework to ensure a fair, unbiased, and repeatable evaluation and conduct detailed analyses assessing the calibration under distributional changes (e.g., distributional shift and application to out-of-distribution data). Furthermore, by investigating the influence of different detector paradigms, post-processing steps, and suitable choices of metrics, we deliver novel insights into why poor detector calibration emerges. Based on these insights, we are able to improve the calibration of a detector by simply finetuning its last layer.
}




\maketitle

\section{Introduction}\label{sec:intro}
Over the last few years, deep object detection has had an impressive evolution. Architectures such as \textsc{Faster-RCNN}~\citep{ren2015faster} and \textsc{DETR}~\citep{carion2020end} showed remarkable performances on challenging datasets. However, these architectures usually lack in assessing their own uncertainty associated with their predictions~\citep{feng2021review}. Often, the quality of probabilistic predictions is poor and should not be trusted out of the box. For example, consider a detection task in an autonomous driving environment. Here, over- or underconfident predictions leading to wrong decisions may result in accidents with cars or even more vulnerable road users such as cyclists~\citep{bieshaar_cooperative_2018}.

% \begin{figure*}
%     \centering
%     \includegraphics[width=.9\linewidth]{images/graphical_abstract.pdf}
%     \caption{Calibration plot and detector predictions from \textsc{DETR} trained on a subset of COCO.}
%     \label{fig:graphical_abstract}
% \end{figure*}
% \newcommand\factor{.80}
\begin{figure*}
    \centering
    \begin{subfigure}{0.33\linewidth}
        \centering
        \includegraphics[width=\linewidth]{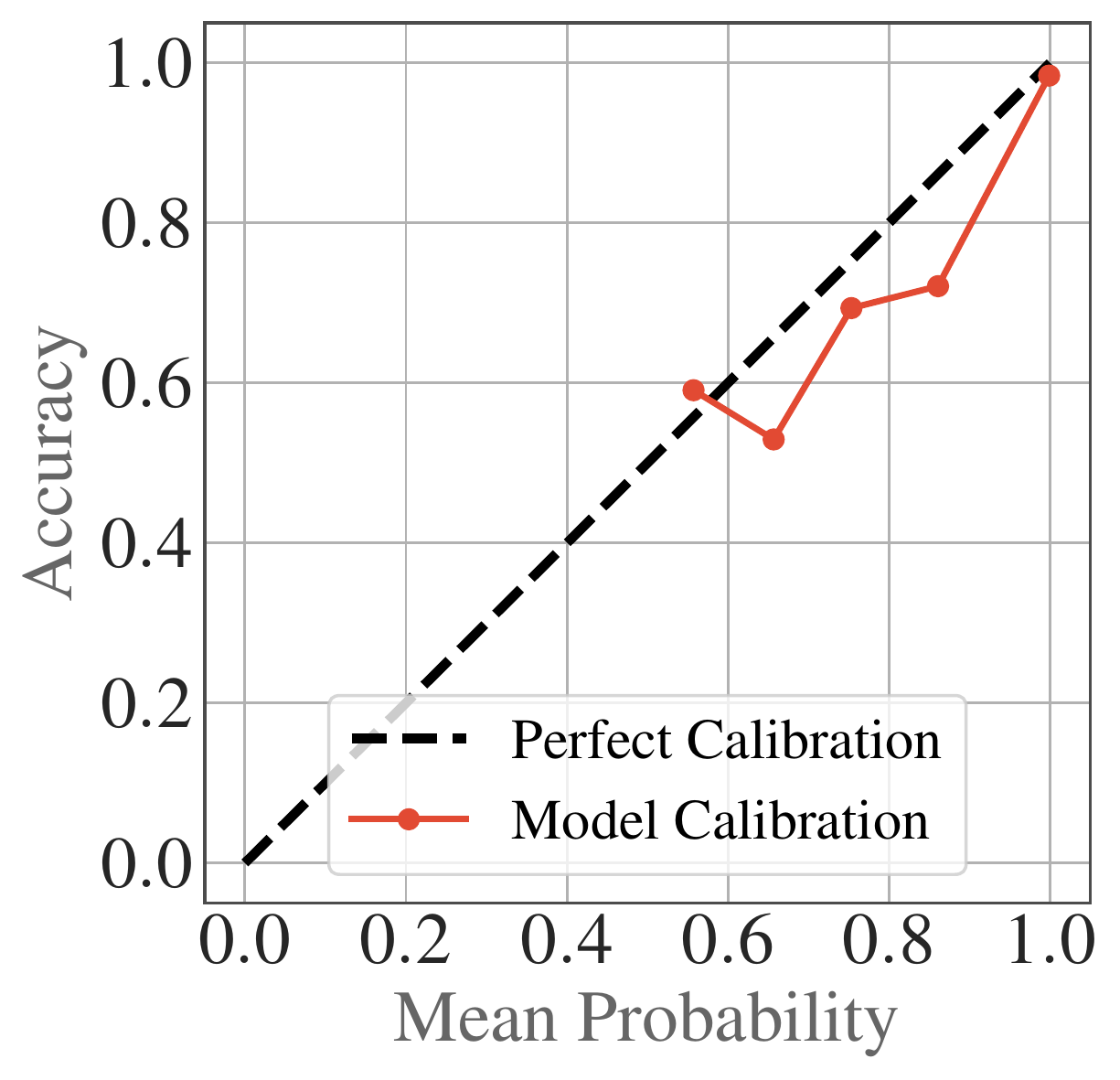}
        \caption{Top-Label Calibration Plot}
        \label{fig:graphical_abstract:a}
    \end{subfigure}%
    \begin{subfigure}{.33\linewidth}
        \centering
        \includegraphics[width=\linewidth]{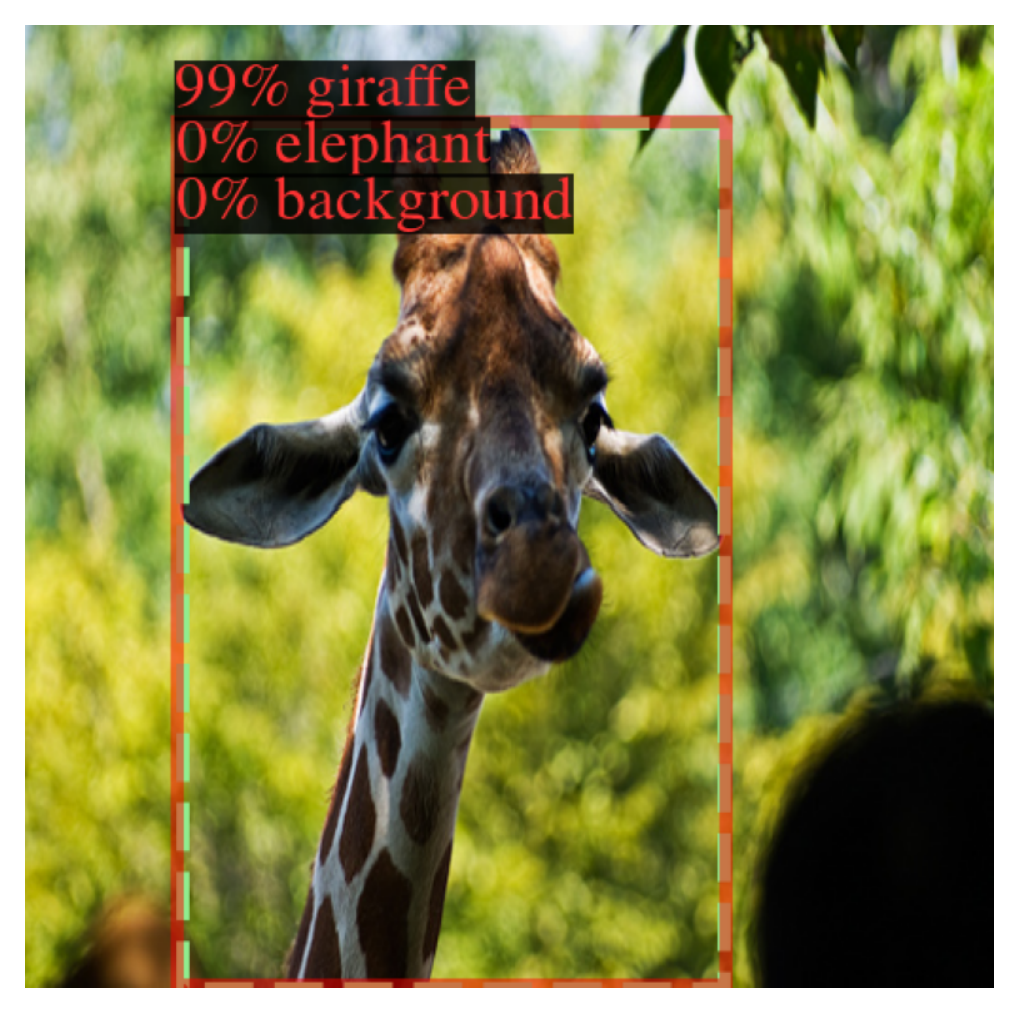}
        \caption{Confident Prediction}
        \label{fig:graphical_abstract:b}
    \end{subfigure}%
    \begin{subfigure}{.33\linewidth}
        \centering
        \includegraphics[width=\linewidth]{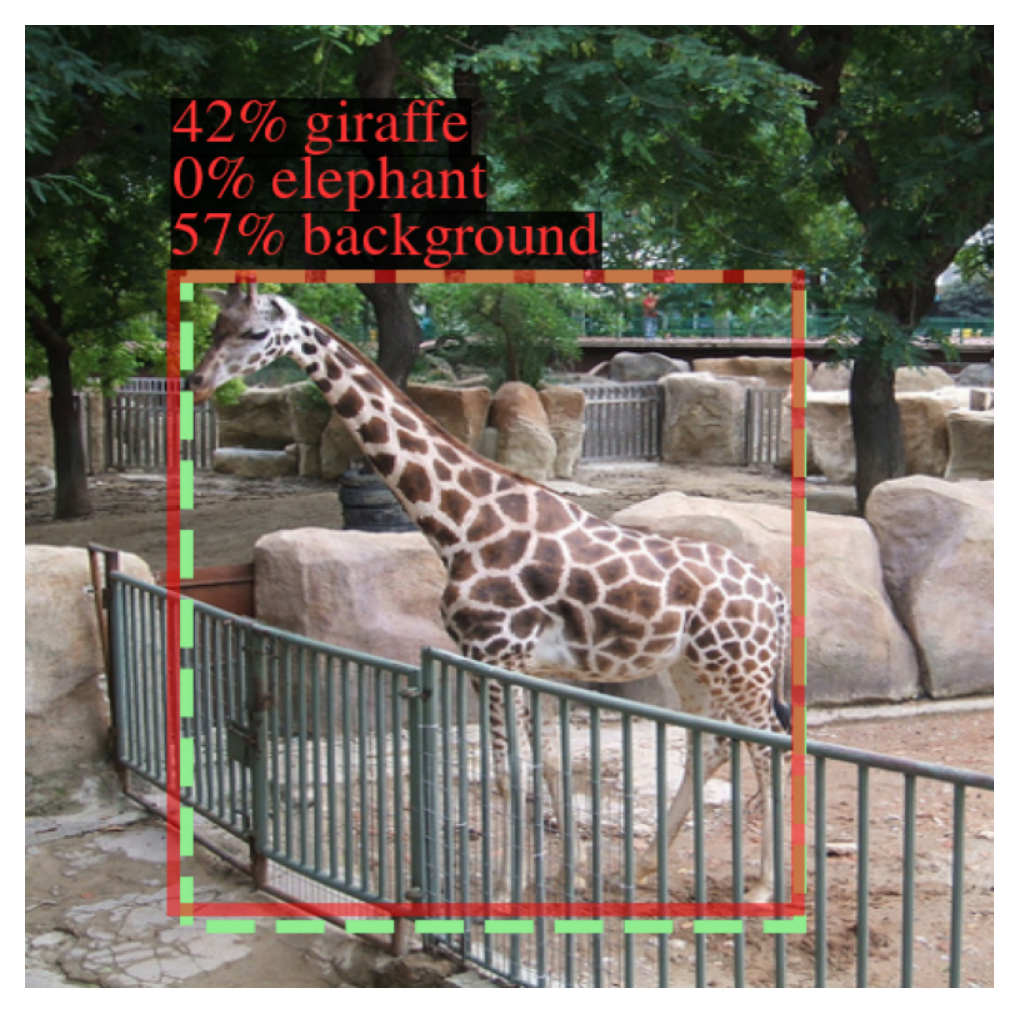}
        \caption{Uncertain Prediction}
        \label{fig:graphical_abstract:c}
    \end{subfigure}%
    \caption{Calibration plot and detector predictions from \textsc{DETR} trained on a subset of COCO.}
    \label{fig:graphical_abstract}
\end{figure*}
% Properties of a good DNN: Calibration + OOD Detection
Ideally, we wish to have a detector capable of providing well-calibrated probabilistic predictions while also being able to recognize object-like entities in images that do not originate from the training data distribution. In the case of well-calibrated probabilistic predictions such as class probabilities, we would like a probability to represent the actual occurrence frequency. For instance, if our detector predicts ``pedestrian'' with a probability of $0.8$, then we expect the ground truth to be a pedestrian $80$ percent of the time. The concept of how well a model can reflect this is referred to as \emph{uncertainty calibration}~\citep{guo2017calibration}. The latter property enables to identify distributional changes, which is crucial when deploying detectors in the real world, especially if the detector is faced data samples that are not well covered by the training data distribution~\citep{ovadia2019can}. Thereby, we can determine to which degree we can trust a prediction. Consider a scenario from autonomous driving in which a flock of sheep crosses a road and our detector was only trained on cars and pedestrians. In such a case, the network should correctly identify this problematic situation due to a distributional change of the samples and leave further decisions to the driver. 

{
Figure~\ref{fig:graphical_abstract} illustrates the attempt of assessing the calibration quality for the popular pretrained object detector \textsc{DETR} finetuned on a subset of the COCO dataset~\citep{lin2014microsoft} with two object classes (giraffe and elephant). 
Figure~\ref{fig:graphical_abstract}~(a) represents the \textit{Top-Label Calibration Plot} (TCP) that only considers the highest predicted probabilities of a detector as done in the literature (e.g., expected calibration error in~\citep{neumann2018relaxed}). As these should reflect the actual occurrence, this diagram shows accuracy as a function of predicted probability. A perfectly calibrated detector would yield a model calibration (red) lying on the diagonal (black), with the predicted probability equal to the number of correct predictions. Hence, in this example, we seem to observe a well-calibrated detector. Note, the red curve starts at about 0.5 because there are no lower maximum probabilities.
In Fig.~\ref{fig:graphical_abstract}~(b) and Fig.~\ref{fig:graphical_abstract}~(c), we see detector predictions on two images. The actual predictions are marked red, while the ground truth boxes are marked green. Figure~\ref{fig:graphical_abstract}~(b) demonstrates a very confident but reasonable prediction for the class giraffe. Accordingly, this prediction will positively influence the assessment in the calibration plot on the left. In contrast, the right figure shows an uncertain false prediction for the background class presumably because of the fence. 
Here, the probability for the class background is only slightly higher than the one for the correct class giraffe. This outlines the importance of considering all available information for evaluation (i.e., all probabilities, not only the highest ones). Thus, the TCP should not be used for evaluation. In the remainder of this article, we will also see that we should assess the calibration of different classes individually.
}

\section{Contributions}
% Summary of all contribution in one sentence
In this work, we investigate multi-class calibration properties of different pretrained state-of-the-art object detection architectures without and with distributional changes, i.e., shifted and out-of-distribution data. We provide detailed analyses of two different detector paradigms (i.e., two-stage and set-based) with two architectures (i.e., \textsc{Faster-RCNN} and \textsc{DETR}), examine their ability to identify distributional changes, and their calibration quality from two perspectives.
% perspectives
In the first perspective, we consider the inference of an object detection architecture as suggested by the literature with the aim to achieve a high generalization performance in the form of mean averge precsision (mAP)~\citep{lin2014microsoft}. More precisely, this means that the raw predictions of a detector are often \textit{post-processed} (e.g., removing duplicates) to obtain a more appropriate prediction set for an image. We often encounter such post-processing steps (e.g., non-maximum suppression, NMS) when working with popular object detection architectures such as \textsc{Faster-RCNN}. In the second perspective, we focus on the \textit{raw outputs of the DNN} with the goal to examine the quality of the probabilistic predictions before applying any post-processing steps. Consequently, we focus on assessing the DNN's calibration and not the calibration of the whole detection pipeline. This way, we intend to give deeper insights, hopefully guiding future research and making it easier to develop techniques for uncertainty modeling in object detection.
% multi class
Furthermore, in contrast to related work evaluating only the highest class probability~\citep{schwaiger2021black}, we focus on a \textit{multi-class setting} in which all class probabilities are considered when evaluating the calibration of the detectors. As will be shown in this article, while detectors often appear to be well-calibrated regarding the highest predicted class probability, their probabilistic outputs of the remaining classes are not. This can lead to severe problems in many cases. For instance, suppose a detector's highest predicted probability is $0.6$ for the class pedestrian and its second-highest probability is $0.4$ for the class e-scooter driver. In such an example, it is evident that this probability must also be well-calibrated to avoid wrong decisions. %In many areas, such a behavior can lead to severe problems, e.g., when making predictions in a cost-sensitive setting.
% distribution shift
Moreover, since changes in the \textit{data distribution} are widespread in real applications, it is essential that detectors are able to recognize and handle them~\citep{ovadia2019can}. Accordingly, we evaluate the uncertainty of detectors under distributional shifts (i.e., different sample but approx.~same class distributions) and also their ability to identify out-of-distribution samples (i.e., both sample and class distributions differ).
%, i.e., samples originating from an entirely different distribution. 
More specifically, we construct shifted and out-of-distribution dataset versions to assess the properties of the detectors.
% General framework for unbiased and reproducible evaluation
In summary, we investigate the following research questions:
\begin{itemize}
    \item How can we build a modular calibration evaluation framework that is suitable for various object detection architectures and ensures an unbiased and repeatable evaluation?
    % \item How should we evaluate multi-class object detection architectures regarding their calibration and which metrics should be considered in the evaluation?
    \item Which metrics should be used to evaluate multi-class object detection architectures regarding their calibration and how should these metrics be applied?
    \item How do post-processing steps in a detection pipeline influence the calibration and do architectures that avoid them deliver better calibrated predictions?
    \item How well are the class probabilities of pretrained object detectors calibrated when the detectors are applied on samples from shifted versions of the training distribution?
    \item How well can a detector identify new objects from out-of-distribution data by means of its probabilistic outputs?
\end{itemize}
Based on the experimental results and findings, we conduct a case study in which we re-calibrate an object detector by simply finetuning its last layers through changing the importance of the background class. Our implementation is publicly available at \url{https://github.com/ies-research/uncertainty-object-detection}.

% Structure
The remainder of this article is organized as follows: 
In Sect.~\ref{sec:problem_setting}, we give a formal definition of our problem setting and introduce the considered object detection architectures which will be exemplarily used as they are the most prominent representatives of their paradigms.
Section~\ref{sec:related_work} analyzes related research regarding uncertainty calibration in classification and object detection.
Afterward, in Sect.~\ref{sec:eval_method}, we propose our evaluation framework and the employed metrics, which allow us to assess the calibration and uncertainty of object detection architectures, and in Sect.~\ref{sec:experiments}, we address the aforementioned research questions by quantitative and qualitative analyses. Based on these insights, in Sect.~\ref{sec:recal}, we recalibrate a detector by finetuning its last layers. Finally, in Sect.~\ref{sec:conclusion}, we conclude our work and highlight potential future research directions.  

%The first version leaves the class distribution unchanged and only adjusts the data distribution. In the second version, both the data and class distribution deviate from the training dataset, intended to represent an out-of-distribution scenario.

\section{Problem Setting}\label{sec:problem_setting}
This section introduces the notation regarding object detection and calibration used throughout this article and the architectures we evaluate.

\subsection{Notation}
% problem
In our setting, we consider object detection problems for computer vision. We represent a color image (i.e., input sample to an object detector) by a tensor $\x \in \set{X}$, where $\set{X} = \mathbb{R}^{W \times H \times C}$ describes the space of all possible images with $H, W, C \in \mathbb{N}$ as height, width, and number of color channels.. An image $\x$ can contain an unknown number of objects, of which each is represented by a box $\b \in \set{B}$ describing its position and a label $\y \in \set{Y}$ explaining its class. The sets $\set{B} = [0, 1]^4$ and $\set{Y} = \{1, \dots, K\}$ define the space of all possible boxes and $K \in \mathbb{N}_{>1}$ class labels, respectively. We define the target set for a single image as $\set{T} \in \set{P}(\set{Y}\times\set{B})$ where $\set{P}(\cdot)$ denotes the power set. For example, an image $\x_n$ with two objects would have the target set $\set{T}_n = \{(y_1, \vec{b}_1), (y_2, \vec{b}_2)\}$. Finally, we describe a dataset consisting of images and targets as $\set{D} = \{(\vec{x}_n, \set{T}_n)\}_{n=1}^N$ where we have a total number of $N \in \mathbb{N}_{\gt 0}$ images. The images $\vec{x}_n$ are distributed according to the distribution $p(\vec{x})$ and boxes $\vec{b}_{nt}$ and labels $y_{nt}$ in the target set $\set{T}_n$ are assumed to be distributed according to $p(\vec{b} \vert \vec{x}_n)$ and $ p(y \vert \vec{b}_{nt}, \vec{x}_n)$, respectively.

% detector model
Formally, an object detector is a function $\detector: \set{X} \to \mathcal{P}(\simplex{K} \times \mathcal{B})$, where $\detectorParameters$ is the set of trainable parameters usually optimized with techniques such as gradient descent~\citep{bishop2006pattern} and $\simplex{K}$ is the $K$-simplex within the $K+1$ dimensional unit hypercube spanned by the $K$ classes and the additional background class $\background$~\citep{ren2015faster,carion2020end}. Most object detectors employ a background class to distinguish whether an object is present in a proposed region. Hence, an object detector is a function that takes an image $\x_n$ as input and outputs a set of predictions $\hat{\set{T}}_n = \{(\hat{\p}_{nt}, \hat{\b}_{nt})\}_{t=1}^{\vert\hat{\set{T}}_n\vert}$ where $\hat{\p}_n \in \simplex{K}$ are class probabilities for the corresponding box $\hat{\b}_n~\in~\set{B}$. The number of predictions per image $\vert\hat{\set{T}}_n\vert \in \mathbb{N}$ might vary depending on the architecture. % and, in practice, this is frequently realized by specifying an upper limit of predictions (e.g., $\vert\hat{\set{T}}_n\vert = 100$)~\citep{carion_end--end_2020}.% and filtering out those with the highest probability for the background class~$\background$~\cite{detr}.

% calibration
%In contrast to~\cite{harakeh,feng} explicitly considering regression uncertainties, we focus on unaltered object detection models and thus on assessing the calibration of class distributions.
Calibration expresses the quality of the predicted probabilities of a model. Formally, for all probability vectors $\hat{\p}$ of a trained detector on the simplex $\simplex{K}$, we want to satisfy
\begin{align}
    P(Y=y \vert \hat{\p}) = \hat{p}_y \text{ for all } y \in \set{Y}\cup\{\background\},\label{eq:calibration}
\end{align}
where $Y$ is the random variable for the true class.
Intuitively, for an object detector, this means that all predicted probabilities $\hat{p}_y$ for a prediction $(\hat{\p}, \hat{\b})$ should match their true (but unknown) probability. For example, collecting all predictions where the object detector returned a probability of $0.2$ for class pedestrian, we want $20$\% of them to actually be a pedestrian. Furthermore, this should hold for all classes $y\in\set{Y}$ and probabilities $p_y \in [0, 1]$. In contrast to similar works which assess the calibration based solely on the predicted class probability $\max \hat{\p}$~\citep{kuppers2020multivariate,neumann2018relaxed}, we focus on calibration always considering all class probabilities as suggested in~\cite{kumar2019verified}.
% Note, that there exists no model which is perfectly calibrated and can satisfy the condition in Eq.~\eqref{eq:calibration}. 

% Recalibration - weg da scope von paper sich geändert hat
%Since probabilities of deep learning models are poorly calibrated, we often consider recalibration methods for improvement~\cite{kumar2019verified}. In particular, in this work we focus on post-hoc calibration, where we exploit a calibration dataset~$\set{D}_{\mathrm{Cal}}$ and an additional calibration model~$\calibrator$. This calibration model usually transforms our model's predicted probabilities into calibrated ones. Similar to~\cite{kuppers2020multivariate}, we employ box-sensitive calibration models $\calibrator: \simplex{K} \times \set{B} \to \simplex{K}$, which also consider the object's position by using a detection $(\hat{\p}, \hat{\b})$ as its input.

\subsection{Detection Architectures}
This work examines two essential kinds of object detection paradigms, namely \textit{two-stage} and \textit{set-based}. Specifically, we focus on the architectures \textsc{Faster-RCNN} and \textsc{DETR}, as they represent the paradigms' most prominent representatives. Since architectures of the \textit{one-stage} paradigm do not directly output multi-class probability vectors as defined in our problem setting, we leave its investigation for future work. %They have the property that the output of the classification head $\hat{\vec{p}} \in \simplex{K}$ is interpreted as the parameters of a categorical distribution $\text{Cat}(y\vert\hat{\vec{p}})$. 
In the following, we briefly describe how the architectures and their detection pipeline work, including the respective post-processing steps.

\textsc{Faster-RCNN}~\citep{ren2015faster} is one of the most prominent object detection architectures and its forward-propagation is performed in two stages. First, it uses a so-called region proposal network (RPN) to predict multiple regions that may contain potential objects. The second stage utilizes these regions and deploys another DNN to solve a simple classification and regression problem. The output of this network is called prediction and is denoted by $\detector(\vec{x})$. In this article, we only focus on the calibration of the second stage and leave the classification problem in the RPN (first stage) for future work. 
For specifying the post-processing steps, we consider the implementation of detectron2~\citep{wu_detectron2_2019}. Out of $1000$ predictions, we discard those where the maximum probability within $\hat{\vec{p}}$ is below a certain threshold. Additionally, to avoid duplicate predictions for the same ground truth object, we filter them out using NMS. That means, that we have a varying number of predictions per image $\vert\hat{\set{T}}_n\vert$.

\textsc{DETR}~\citep{carion2020end} is a set-based object detection architecture based on transformers~\cite{vaswani2017attention}. The term set-based arises as \textsc{DETR} views object detection as a direct set prediction problem in which we can detect objects directly without the need for anchors or region proposals. To achieve this, \textsc{DETR} learns attention weights in a transformer describing the pixel and object relationships in an image. In contrast to other object detection architectures, \textsc{DETR} learns to avoid duplicates through its set-based loss function and therefore does not require any post-processing steps.
This means, that it predicts a fixed number of $\vert\hat{\set{T}}_n\vert = 100$ objects per image, as suggested by the authors.

\section{Related Work}\label{sec:related_work}
% Motivation and Problems: Object Detection and Uncertainty 
Commonly, calibration properties are modeled and researched in the context of image classification problems~\citep{ovadia2019can}, and in recent years, considerable work has been done in that field. 

For image classification, \cite{guo2017calibration} showed that although modern architectures such as ResNet~\citep{he2016deep} achieve outstanding generalization performance, they provide poorly calibrated and overconfident outputs. They introduced temperature scaling, a multi-class extension to Platt scaling, which improves a DNN's calibration by scaling its predictions with a parameter learned from a separate dataset. \cite{kull2019beyond} noticed that, while temperature scaling improves the calibration of the highest predicted probability of a vector containing all class probabilities, the remaining probabilities are still poorly calibrated.  Accordingly, they proposed Dirichlet Calibration for the multi-class setting to improve the quality of all predicted probabilities. Furthermore, in a large-scale evaluation, \cite{ovadia2019can} assessed the behavior of various models under distributional changes and showed a deteriorated quality of their uncertainty estimates. 

For classification problems, we only need to consider that the input of a model is assigned to a single class. In object detection, however, we need to consider an unknown number of objects per image, jointly solve regression and classification problems, and often use heuristics such as NMS to obtain our final predictions. These complexities make the transfer of existing uncertainty modeling and calibration evaluation concepts to the object detection task challenging. To the best of our knowledge, one of the first methods regarding uncertainty calibration in deep object detection was proposed by \cite{neumann2018relaxed}. Their article demonstrates the poor calibration of pedestrian detection models and proposes an extension of temperature scaling, avoiding the need for a separate calibration dataset. More recently, \cite{kuppers2020multivariate} discovered that the calibration of a model depends on an object's position in an image and proposed box-sensitive recalibration methods for improvement. As most works focus on the evaluation of the entire detection pipeline, \cite{schwaiger2021black} assessed the influence of NMS on the calibration. They examined the highest predicted probability of DNNs used in different detectors and showed the negative impact of NMS on the calibration quality.

In summary, the related work regarding calibration in object detection is sparse, often focuses only on the highest probability leading to a loss in information in multi-class settings, and frequently employs inappropriate metrics for evaluation.
In this article, we focus on these aspects regarding the evaluation of pretrained detectors. An in-depth study on recalibration techniques of detectors is subject to our future work.

\section{Evaluation Framework}\label{sec:eval_method}
\begin{figure*}
    \centering
    \includegraphics[width=\textwidth]{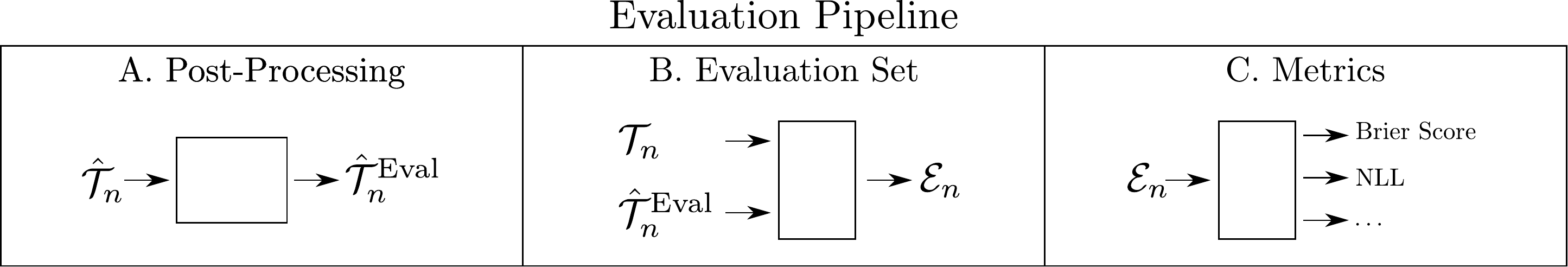}
    \caption{Our modular evaluation framework for each architecture consists of three steps.}
    \label{fig:eval_framework}
\end{figure*}
The literature suggests many techniques for evaluating the calibration and uncertainty of object detection architectures. However, most of these do not clearly define the steps toward an evaluation metric. In particular, it is not clearly described how to incorporate the predictions of object classes (i.e., true and false positives, TP and FP), the predictions of the background class (i.e., true negatives, TN) and missing predictions (i.e., false negatives, FN). \textit{As an attempt to answer our first research question and to make the evaluation understandable and repeatable, we propose a modular evaluation pipeline consisting of three steps} (cf.~Fig.~\ref{fig:eval_framework}). Starting point for the evaluation is the raw output $\detector(\vec{x})$ of an object detectors's forward propagation. To obtain an evaluation metric, we look at the subsequent three steps:
\begin{enumerate}
    \item Post-Processing: Involves filtering out redundant predictions as done by NMS.
    \item Creation of an Evaluation Set: Involves matching predictions with ground truth objects.
    \item Metric Evaluation: Involves defining calibration metrics based on the evaluation set.
\end{enumerate}
Each of these steps can be adapted according to the requirements of a specific architecture. The upcoming subsections highlight essential design choices in each step that are necessary for a fair evaluation of all considered architectures.
%the remainder of this paper, we will investigate our detectors from two perspectives.
%In an \textit{application-based perspective}, we evaluate the detections obtained by following the post-processing steps of the respective architecture (e.g., NMS in Faster-RCNN). In a \textit{modeling-based perspective}, we evaluate all detections of the detector and avoid heuristic post-processing steps that may bias the evaluation of the model.

%This section compares our method against uncalibrated object detectors as well as some baselines. Therefore, we first transfer common calibration evaluation techniques from classification to our problem setting. Based on this, we reformulate popular proper scoring rules~\cite{} and extend the concept of the calibration error and plots by considering the precision of a detector. Other evaluation setups~\cite{feng2021review} employ a confidence threshold to filter low confidence predictions. In contrast, we consider all predictions as we intend to evaluate all outputted probabilities of a detector and not only those above a certain threshold. Furthermore, be aware that we do not assess missing detections (i.e.,~false negatives) as we are only able to evaluate probabilities if a prediction is present. This also implies that it is important to assess the subsequent metrics only in combination with the detector's performance (i.e.,~mAP).

\subsection{Post-Processing}
% prediction set and filtering detections
Since there exist a wide range of object detection paradigms and architectures, there are also a lot of different detector-specific ways to obtain a final prediction set for an image. Many detection pipelines employ NMS and filter out several predictions based on different criteria to achieve satisfying generalization performances (measured, e.g., with mAP)~\citep{ren2015faster}. However, these post-processing steps remove predictions from the predictions set $\hat{\set{T}}_n$, which might carry valuable information regarding calibration and uncertainty assessment. Thus, it is essential to be aware of specific post-processing steps that a detection pipeline uses as these might bias the evaluation, especially when evaluating uncertainties~\citep{schwaiger2021black}.
In our evaluation, we specify a set $\hat{\set{T}}_n^{\text{Eval}}$ to define the (final) prediction set for sample $\x_n$ that are considered during evaluation. For example, when considering NMS as a post-processing method, this set will be a subset of the original predictions, i.e.,~$\hat{\set{T}}_n^{\text{Eval}} \subseteq \hat{\set{T}}_n$.

On one hand, we consider this set from a perspective which treats a detector as a black box. In particular, we use post-processing steps as it would be done when deploying the detection pipeline in practice. 
For example, for the detection architecture \textsc{Faster-RCNN}, the predictions $\hat{\set{T}}_n^{\text{Eval}}$ for image $\x_n$ after post-processing are defined by i)~removing all predictions $(\hat{\vec{p}}_{nt}, \hat{\vec{b}}_{nt})$ where the maximum of $\hat{\vec{p}}_{nt}$ is below a certain threshold, ii)~removing all duplicate predictions with NMS, and iii)~keeping a certain number of predictions with the highest maximum probability. At this point, it becomes apparent that with this perspective we will neglect many predictions in the subsequent course of the evaluation. These might have better calibrated class probabilities or we might lose potentially helpful predictions which we could exploit to identify distributional changes.

On the other hand, we also intend to evaluate a detector in a more detailed perspective without post-processing. Specifically, we determine the calibration quality of $\detector$ and not of the entire detection pipeline by considering the raw predictions that were made by the DNN (i.e., $\detector$) without filtering any predictions. Thus, our set for evaluation $\hat{\set{T}}_n^{\text{Eval}}$ is given by all the predictions $\hat{\set{T}}_n$ of the DNN.
% Using that perspective,  as done in the first perspective.
% In contrast to~\cite{feng2021review, whitebox}, we do not consider a confidence thresholds as this would filter low confidence detection which we wish to evaluate. %Furthermore, as our considered detector model is DETR we do not need to worry about the use of NMS.

\subsection{Evaluation Set}
% Assignment of gt to pred and the right type of assignments
In order to be able to evaluate the predictions, it is necessary to have a suitable assignment of predictions from the prediction set $\hat{\set{T}}^{\text{Eval}}_n$ to ground-truth objects in the target set $\set{T}_n$. We can realize such an assignment in multiple ways. However, it is important to note that the previously mentioned perspectives leading to our prediction set $\hat{\set{T}}^{\text{Eval}}_n$ play a critical role in that selection. 
Generally, if the detector aims at avoiding duplicates during inference (e.g.,~\textsc{Faster-RCNN} with post-processing), then we have to force an assignment of a single prediction to a single ground truth object. Otherwise, we would not penalize duplicate predictions (FP) in this case. 
Conversely, if our detector predicts duplicates (e.g.,~\textsc{Faster-RCNN} without post-processing), we must ensure that multiple predictions can be assigned to a single ground truth object. Otherwise, we would penalize duplicate predictions, in this case, even if the detector is supposed to make multiple per object.
As an example, consider evaluating the raw predictions of \textsc{Faster-RCNN}. Here, we must ensure that we assign multiple predictions to a single ground truth object. \textsc{Faster-RCNN} without NMS may correctly predict multiple boxes for a single ground truth object. On the other hand, it is essential to use a one-to-one assignment for \textsc{DETR}, since we expect that a \textsc{DETR} based detector learned to avoid duplicate predictions during inference.

% Definition of our assignment and definition of Evaluation set
To build our evaluation set $\set{E}_n$ for a sample, we therefore focus on a simple matching of predictions to ground truth objects based on the \textit{Intersection over Union} (IOU) between predicted and ground truth box. In particular, anything above an IOU threshold of $0.5$ is considered as matched. We match the pair with the highest IOU when we need a one-to-one assignment. Note that we prefer this simple matching over the matching used in the respective architecture (e.g.,~\textsc{DETR}'s Hungarian matcher) to compare different architectures against each other in the same evaluation setting.
We define the index sets of matches based on the predictions $\hat{\set{T}}^{\text{Eval}}_n$ and targets $\set{T}_n$ as
\newcommand{\BS}{\operatorname{BS}}
\newcommand{\NLL}{\operatorname{NLL}}
\newcommand{\TCE}{\operatorname{TCE}}
\newcommand{\MCE}{\operatorname{MCE}}
\newcommand{\dTCE}{\operatorname{dTCE}}
\newcommand{\dMCE}{\operatorname{dMCE}}
\newcommand{\E}{\set{E}}
\newcommand{\M}{\set{M}_n}
\newcommand{\Mpred}{\overline{\set{M}}^{\set{T}}_n}
\newcommand{\Mgt}{\overline{\set{M}}^{\hat{\set{T}}}_n}
\newcommand{\yBackground}{y^\mathrm{\otimes}}
\newcommand{\bBackground}{\b^\mathrm{\otimes}}
\newcommand{\pBackground}{\hat{\p}^{\mathrm{\otimes}}}
\begin{align}
    \M &= \{ (i,j) \mid (y_{ni}, \vec{b}_{ni})  \text{ matches } (\hat{\vec{p}}_{nj}, \hat{\vec{b}}_{nj}) \}, \\
    \Mpred &= \{ j \mid   (\hat{\vec{p}}_{nj}, \hat{\vec{b}}_{nj}) \text{ is not matched}\}, \\
    \Mgt &= \{ i \mid (y_{ni}, \vec{b}_{ni}) \text{ is not matched} \}. 
\end{align}
Consequently, for our evaluation, we define the evaluation set as
\newcommand{\textunderbrace}[2]{{%
  \underbrace{#1}_{\text{#2}}
}}
\begin{align}
\begin{gathered}
    \E_n = \underbrace{\{(y_{ni}, \b_{ni}, \hat{\vec{p}}_{nj}, \hat{\vec{b}}_{nj})\}_{(i, j) \in \M}}_{\text{\makebox[0pt]{matched predictions (TP and FP)}}} \\
    \cup \underbrace{\{(\yBackground, \bBackground, \hat{\vec{p}}_{nj}, \hat{\vec{b}}_{nj})\}_{j \in \Mpred}}_{\text{\makebox[0pt]{unmatched predictions (FP and TN)}}}\\
    \cup \underbrace{\{(y_{ni}, \b_{ni}, \pBackground, \bBackground)\}_{i \in \Mgt}}_{\text{\makebox[0pt]{missing predictions (FN)}}},\label{eq:eval_set}
\end{gathered}
\end{align}
where $\yBackground = K+1$ denotes the label for the background class, $\bBackground \in \set{B}$ is an arbitrary box, and $\pBackground \in \simplex{K}$ is a probability vector assigning all its probability mass to the background class $K+1$. With this set, our evaluation considers not only all predictions (TP, FP, and TN) but also missing ones (FN). Hence, we obtain an unbiased evaluation as we do not ignore potentially valuable predictions.

\subsection{Metric Definition}
\renewcommand{\b}{\vec{b}}
\newcommand{\bHat}{\hat{\b}}
\renewcommand{\p}{\vec{p}}
\newcommand{\pHat}{\hat{\p}}
\renewcommand{\det}{(\pHat_{nt}, \bHat_{nt})}
\newcommand{\acc}{\operatorname{acc}}
\newcommand{\conf}{\operatorname{prob}}
\renewcommand{\B}{\set{B}}
\newcommand{\yonehot}{\vec{y}}
\newcommand{\indicator}[1]{\mathbb{I}(#1)}

The final step toward our framework consists of defining the required metrics based on the evaluation sets $\E_n$. More specifically, we focus on proper scoring rules~\citep{feng2021review}, calibration plots and errors~\citep{kumar2019verified} for the evaluation of the calibration, and on the entropy~\citep{ovadia2019can} to assess if a detector can identify distributional changes. Note that we can easily incorporate additional metrics for evaluation if necessary.
To further simplify the notation, we define the evaluation set containing all predictions as $\E = \bigcup_{n=1}^N\E_n$, an element of that set as $d \in \E$ (i.e., matched, unmatched, or missing predictions), and a one-hot encoded version for $y_{nt}$ as $\yonehot_{nt}$.

\textbf{Proper scoring rules} are prevalent metrics for evaluating predictive probability distributions. Scoring rules are functions that map a probability distribution $\pHat_{nt}$ and a ground truth label $y_{nt}$ to a score. This score expresses how well distribution $\pHat_{nt}$ matches the ground truth distribution of $y_{nt}$.  Furthermore, we call it proper if only the ground truth distribution leads to a minimum score value.
For our evaluation, we use the \textit{Negative Log-Likelihood} (NLL) defined as
\begin{align}
    \NLL(\E) = -\frac{1}{\vert\E\vert}\sum_{d \in \E}  \vec{y}_{nt}^\mathrm{T} \ln \pHat_{nt},
\end{align}
where $\ln$ is applied element-wise to the entries of $\pHat_{nt}$.
Furthermore, we also use the \textit{Brier Score} (BS) defined as
\begin{align}
    %\BS(\E) = \frac{1}{\sum_{n=1}^{N} \vert\E\vert}\sum_{n=1}^{N} \sum_{d \in \E} \sum_{k=1}^{K} (\hat{p}_{ntk} - y_{ntk})^2
    % \BS(\E) = \frac{1}{\vert\E\vert} \sum_{d \in \E} \sum_{k=1}^{K} (y_{ntk}- \hat{p}_{ntk} )^2
    \BS(\E) = \frac{1}{\vert\E\vert} \sum_{d \in \E} \|\vec{y}_{nt} - \pHat_{nt} \|^2,
\end{align}
where $\|\cdot\|^2$ is the Euclidean distance.
Both the NLL and BS are proper scoring rules and we refer the reader to~\cite{feng2021review} for a more detailed explanation.

\textbf{Calibration plots} or reliability diagrams are often employed to assess a model's calibration visually~\citep{guo2017calibration}. Their idea is to visualize Eq.~\ref{eq:calibration} by plotting approximations of the true (but unknown) probability against the detector's predicted probability. Hence, we estimate both quantities empirically by splitting the probability space $[0, 1]$ into $M$ adjacent bins (i.e., intervals) of equal sizes, assigning all predictions based on their probability, and computing accuracy and mean probability as defined below. On the one hand, we employ the \textit{Top-Label Calibration Plot} (TCP), in which we consider a bin $m$ containing all predictions $\B_m \subseteq \E$ where the maximum probability falls into the respective bin interval.
The mean probability and accuracy are therefore defined by
\begin{align}
    \conf(\B_m) &= \frac{1}{\vert\B_m\vert} \sum_{d \in \B_m} \max\pHat_{nt},\label{eq:tce_conf} \\
    \acc(\B_m) &= \frac{1}{\vert\B_m\vert} \sum_{d \in \B_m} \indicator{\arg\max\pHat_{nt} = y_{nt}}\label{eq:tce_acc},
\end{align}
where $\indicator{\cdot}$ is the indicator function.
Since the TCP only considers the calibration of the most probable class, it neglects to assess all other predicted probabilities. For this reason, we also employ the \textit{Marginal Calibration Plot} (MCP), which depicts the mismatch of all probabilities~\citep{kumar2019verified}.
In such a case, we define $\B_m$ for each class~$k$ individually and only consider the probabilities of the respective class when assigning predictions to a bin. Thus, the average mean probability and accuracy in bin $m$ for class $k$ is defined as
\begin{align}
    \conf_k(\B_m) &= \frac{1}{\vert\B_m\vert} \sum_{d \in \B_m} \hat{p}_{ntk},\label{eq:mce_conf}\\
    \acc_k(\B_m) &= \frac{1}{\vert\B_m\vert} \sum_{d \in \B_m} \indicator{k = y_{nt}}\label{eq:mce_acc}.
\end{align}
By plotting the accuracy (y-axis) against mean probability (x-axis) in a two-dimensional Cartesian coordinate system, we can observe the calibration qualities. A diagonal corresponds to a perfect calibration. In contrast, our detector is overconfident if the line is below the diagonal and underconfident if it is above. Considering the TCP, we only examine the highest predicted probabilities. This results in a single calibration line. In the case of multiple classes, we will have one line per class. As we will see in Sect.~\ref{sec:experiments}, it is crucial not to average the respective metrics per bin since this might bias the curves towards the diagonal.

\textbf{Calibration errors} are aggregated values derived from calibration plots, allowing us to simplify the comparison of different models against each other. Correspondingly, we employ Eq.~\ref{eq:tce_conf} and Eq.~\ref{eq:tce_acc} to define the \textit{Top-Label Calibration Error} (TCE) as
\begin{equation}
\begin{gathered}
    \TCE(\E) = \\\left(\sum_{m=1}^{M}\frac{\vert\B_m\vert}{\vert\E\vert} (\acc(\B_m) - \conf(\B_m))^2\right)^{1/2}.
\end{gathered}
\end{equation}
The TCE is also referred to as the \emph{Expected Calibration Error} in literature~\citep{guo2017calibration}.
Similarly, we use Eq.~\ref{eq:mce_conf} and Eq.~\ref{eq:mce_acc} to define the \textit{Marginal Calibration Error} (MCE) as
\begin{equation}
\begin{gathered}
    \MCE(\E) = \\\left(\sum_{k=1}^{K+1}\sum_{m=1}^{M}\frac{\vert\B_m\vert}{\vert\E\vert} (\acc_k(\B_m) - \conf_k(\B_m))^2\right)^{1/2}.
\end{gathered}
\end{equation}
In addition, we also report the recently proposed detection variants of TCP/TCE and MCP/MCE, which we refer to as dTCP/dTCE and dTCP/dMCE~\citep{kuppers2020multivariate}. Instead of the accuracy, they use the precision as an estimate for the true probability. We utilize an adjusted version of those metrics ignoring box positions to be able to compare them to the other plots and errors. 

\renewcommand{\H}{\operatorname{H}}
\textbf{Entropy} is a measure which can be interpreted as the uncertainty of a distribution and is defined by 
\begin{align}
   \H(\pHat) = - \sum_{k=1}^{K+1} \hat{p}_k \ln \hat{p}_k.
\end{align}
Similar to~\cite{ovadia2019can}, we compute it for all predictions in $\E$ and plot a histogram of entropy values to visualize the uncertainty of a model on a whole dataset. This way, we expect to observe differences resulting from data distributional changes. Since we typically have many background predictions with high probability per image, we consider only matched and missing predictions as defined by Eq.~\ref{eq:eval_set}. This is because we intend to assess the uncertainty of predictions that were made for potential objects of unknown classes. We expect that these uncertainties are higher than the ones obtained from the in-distribution objects.
Note, the entropy is the only metric we can apply to out-of-distribution data, since all other metrics require the same label space $\set{Y}$.

\section{Experiments}\label{sec:experiments}
In this section, we introduce our experimental setup and discuss the results of the experiments that answer our research questions.

\subsection{Setup}
% Pretrained models -> Finetuned them
% Dataset with subsets
We focus on object detectors pretrained on the COCO dataset~\citep{lin2014microsoft} and will evaluate their capability to provide well-calibrated probabilistic predictions as well as their ability to identify distributional changes.
Besides using the 80 class test split of COCO, we want to be able to represent the calibration plots in a clear and understandable way and to investigate the dependence of calibration quality on the number of classes.
Therefore, we additionally construct two subsets of COCO and fine-tune the last layers of detectors by using their respective standard hyperparameter settings for training. In particular, we use a simple subset called \textsc{animals} consisting of images with the two classes giraffe and elephant, and a slightly more complex subset \textsc{traffic} consisting of the ten classes person, bicycle, car, motorcycle, bus, train, truck, traffic light, fire hydrant, and stop sign. We refer to \textsc{all} when talking about the original 80 class COCO test split. Note, as we consider the additional background class in each detector $\detector$, we have at least a three-class classification problem that needs to be solved.

% Distributional Shift - Erklärung was das ist und was es bedeutet?
Generally, a distributional shift occurs when the test sample distribution no longer matches the training sample distribution. To simulate distributional shifts within our evaluation, we follow the works of~\cite{ovadia2019can} and \cite{harakeh2021estimating} and adapt the distribution $p(\x)$ while keeping the distribution $p(y\vert\x, \b)$ approximately unchanged. More precisely, we construct a new dataset with approximately the same class distribution as our respective training dataset but with a different sample distribution. To realize this, we create two subsets of the Open-Images object detection dataset~\citep{kuznetsova2020open} with the same set of classes as in \textsc{animals} and \textsc{traffic} and evaluate them in an identical setting. Like~\cite{harakeh2021estimating}, we assume that these constructed datasets can be interpreted as a shifted version of the training dataset due to a different data collection process (i.e., image quality, sources, and difficulty). %annotation mechanism).

% OOD - Erklärung was das ist und was es bedeutet?
The evaluation on out-of-distribution (OOD) data is commonly done in classification~\citep{huseljic2021separation} as it gives insights about the quality of uncertainty estimates of a model (e.g., probabilistic outputs or derived measures). The idea behind an OOD evaluation in an object detection setting, however, is fairly uncommon~\citep{du2022vos}. We argue that a detector should be able to identify object-like entities in OOD images while returning high uncertainty for them. For example, consider a flock of sheep crossing a road and a detector in an autonomous vehicle trained on traffic classes such as cars or pedestrians. Without returning highly uncertain predictions for the unknown objects (i.e., sheep), we would not be able to identify this situation and only predict the background class with a high probability. Optimally, in such a scenario, our detector should return high uncertainty for all predicted boxes. We evaluate the detectors on out-of-distribution data similar to~\cite{harakeh2021estimating}. Thus, we require a dataset that has not only a different sample distribution $p(\x)$ but also an unknown class distribution $p(y\vert\x,\b)$. Accordingly, we take a subset from Open Images containing images in which none of the 80 classes from COCO appear.
More details to the corresponding datasets, experimental setup, experiments, additional results as well as the implementation of our framework can be found in our implementation.

\subsection{Results}\label{sec:results}
\begin{figure*}
    \centering
    \includegraphics[width=.8\linewidth]{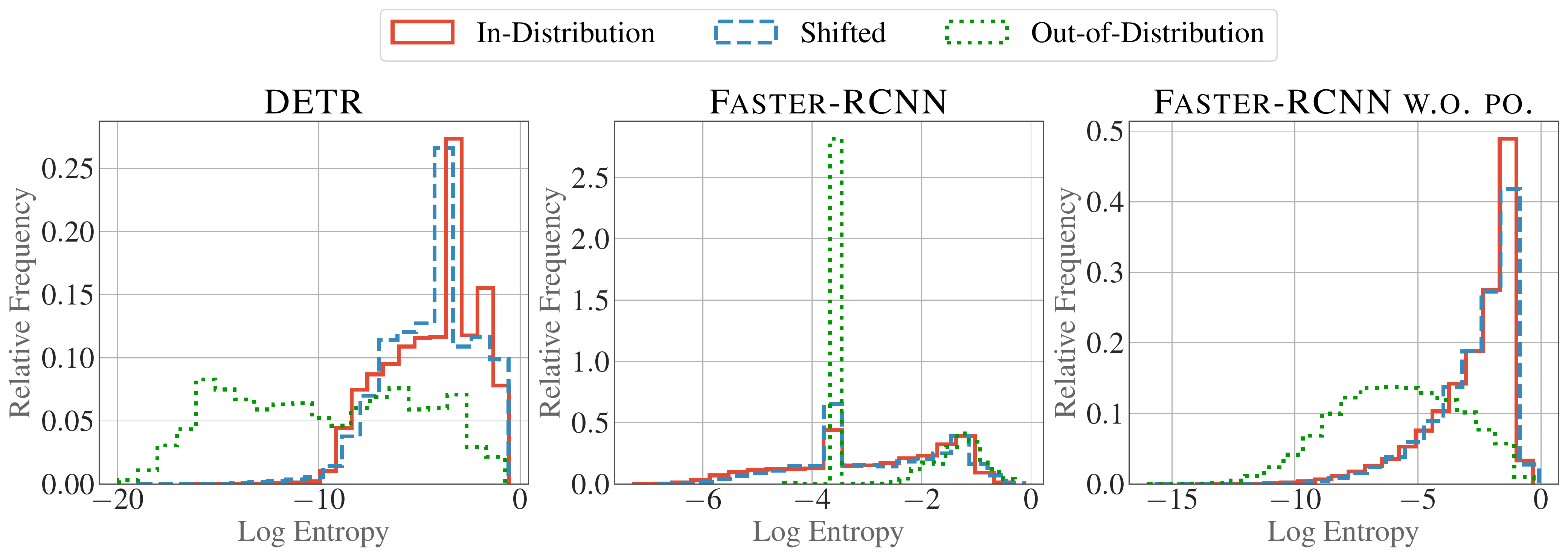}
    \caption{Entropy histograms for in-distribution, shifted, and OOD datasets on the \textsc{traffic} subset.}
    \label{fig:entropy_hists}
\end{figure*}

\renewcommand{\rq}[1]{\textbf{#1}} %research question
\newcommand{\ra}[1]{\textit{#1}} %research answer
In the following, we present the results from numerous conducted experiments to sequentially address our research questions. We report values of all numerical metrics (i.e., generalization and calibration) for \textsc{animals}, \textsc{traffic}, and \textsc{all}, and the corresponding shifted versions in Table~\ref{tab:results}. Additionally, we also show calibration plots (Fig.~\ref{fig:calibration_plots}) for the \textsc{animals} subset and entropy histograms (Fig.~\ref{fig:entropy_hists}) for the OOD subset. We will utilize these to underline the answers to research questions with intuitive explanations. For additional calibration plots, we refer to our implementation.

\begin{table*}[ht!]
\small
\caption{Results from \textsc{DETR} and \textsc{Faster-RCNN} on in-distribution datasets and their respective shifted versions. Arrows next to metrics indicate the direction of the optimal value. We abbreviate \textsc{Faster-RCNN} as \textsc{F-RCNN} and ``without post-processing'' as \textsc{w.o. po.}}
\label{tab:results}
\centering
% \tabcolsep{1}
% Some packages, such as MDW tools, offer better commands for making tables
\begin{tabular}{c|c|c|c|c|c|c|c} % https://tex.stackexchange.com/questions/73283/how-to-use-multirow <---
    \toprule
    \multirow{2}{*}{Dataset} & \multirow{2}{*}{Architecture}  & \multirow{2}{*}{mAP($\uparrow$)} & \multirow{2}{*}{NLL($\downarrow$)} & \multirow{2}{*}{BS($\downarrow$)} & \multirow{2}{*}{TCE($\downarrow$)}  & \multirow{2}{*}{MCE($\downarrow$)} & \multirow{2}{*}{dMCE($\downarrow$)} \\
    {} & {} & {} & {} & {} & {} & {} & {} \\
    \midrule
    \multirow{3}{*}{\shortstack{\textsc{animals} \\(In-Distribution)}}
    & \textsc{DETR}         & \pmb{0.701} & \pmb{0.120} & \pmb{0.036} & \pmb{0.019} & \pmb{0.053} & \pmb{0.046} \\
    & \textsc{F-RCNN}   & 0.639 & 0.549 & 0.268 & 0.061 & 0.140 & 0.122 \\
    & \textsc{F-RCNN w.o.~po.}   & 0.639 & 0.403 & 0.158 & 0.080 & 0.077 & 0.063 \\
    \midrule                                                                     
    \multirow{3}{*}{\shortstack{\textsc{animals} \\(Shifted)}}
    & \textsc{DETR}         & \pmb{0.718} & \pmb{0.105} & \pmb{0.027} & \pmb{0.019} & \pmb{0.046} & \pmb{0.040} \\
    & \textsc{F-RCNN}   & 0.648 & 0.534 & 0.255 & 0.058 & 0.112 & 0.095 \\
    & \textsc{F-RCNN w.o.~po.}   & 0.648 & 0.327 & 0.121 & 0.067 & 0.070 & 0.058 \\
    \midrule\midrule
    \multirow{3}{*}{\shortstack{\textsc{traffic} \\(In-Distribution)}}
    & \textsc{DETR}         & \pmb{0.488} & \pmb{0.209} & \pmb{0.095} & \pmb{0.036} & 0.059 & 0.044 \\
    & \textsc{F-RCNN}   & 0.473 & 0.637 & 0.312 & 0.057 & 0.085 & 0.061 \\
    & \textsc{F-RCNN w.o.~po.}   & 0.473 & 0.360 & 0.155 & 0.065 & \pmb{0.030} & \pmb{0.021} \\

    \midrule                                                                     
    \multirow{3}{*}{\shortstack{\textsc{traffic} \\(Shifted)}}
    & \textsc{DETR}         & 0.381 & 0.371 & 0.163 & 0.067 & 0.099 & 0.073 \\
    & \textsc{F-RCNN}   & 0.387 & 0.878 & 0.434 & 0.135 & 0.150 & 0.108 \\
    & \textsc{F-RCNN w.o.~po.}   & \pmb{0.387} & \pmb{0.335} & \pmb{0.156} & \pmb{0.056} & \pmb{0.046} & \pmb{0.035} \\
    \midrule\midrule
    \multirow{3}{*}{\shortstack{\textsc{all} \\(In-Distribution)}}
    & \textsc{DETR}         &  \pmb{0.420} & \pmb{0.380} & \pmb{0.172} & \pmb{0.046} & 0.030 & 0.021 \\
    & \textsc{F-RCNN}   &  0.392 & 0.792 & 0.364 & 0.087 & 0.036 & 0.023 \\
    & \textsc{F-RCNN w.o.~po.}   &  0.392 & 0.510 & 0.213 & 0.077 & \pmb{0.011} & \pmb{0.006}  \\
    \bottomrule
\end{tabular}
\end{table*}

\newcommand\factor{.80}
\begin{figure*}
    \centering
    \begin{subfigure}{\factor\linewidth}
        \centering
      \includegraphics[width=\linewidth]{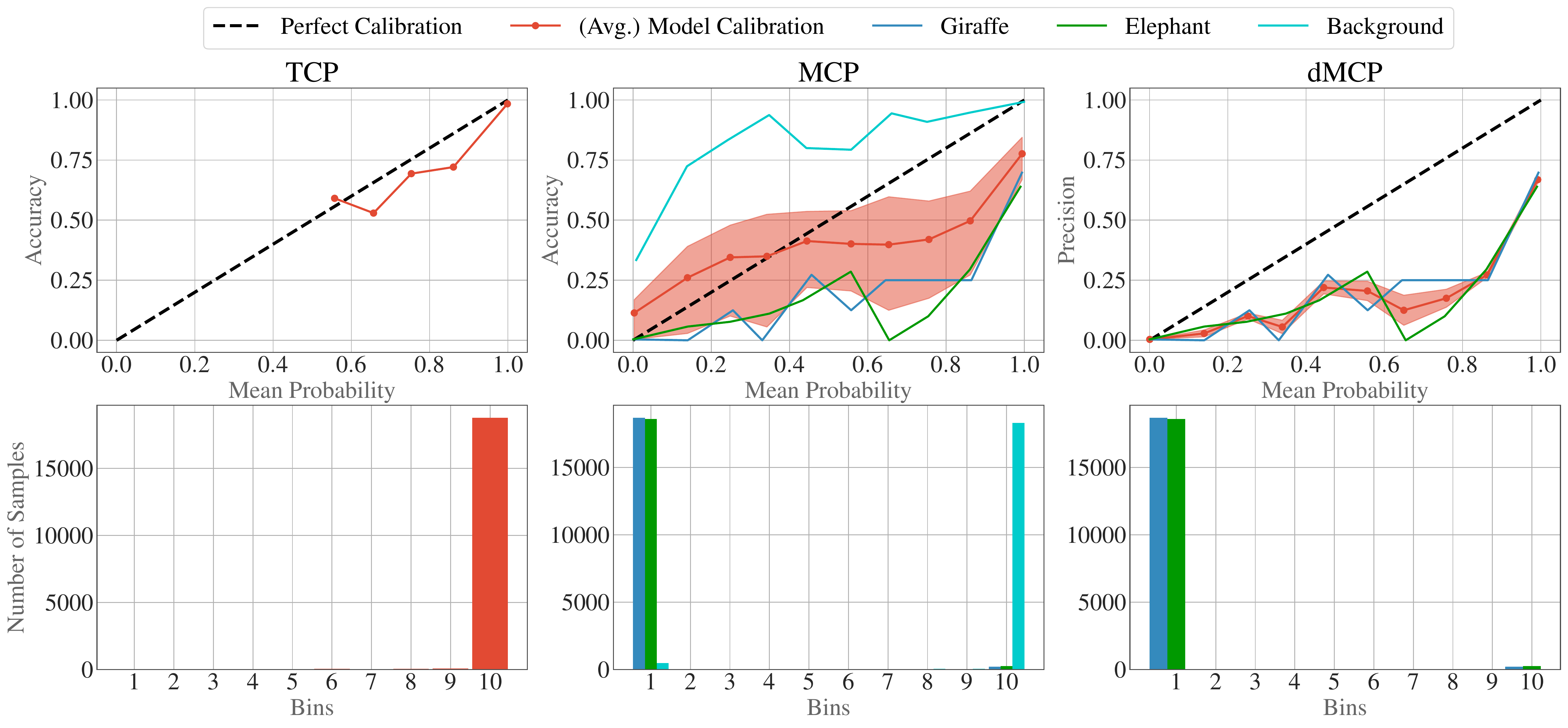}
        \caption{\textsc{DETR} on \textsc{animals} subset.}
        \label{fig:a_detr_animals}
    \end{subfigure}%
    
    \begin{subfigure}{\factor\linewidth}
        \centering
        \includegraphics[width=\linewidth]{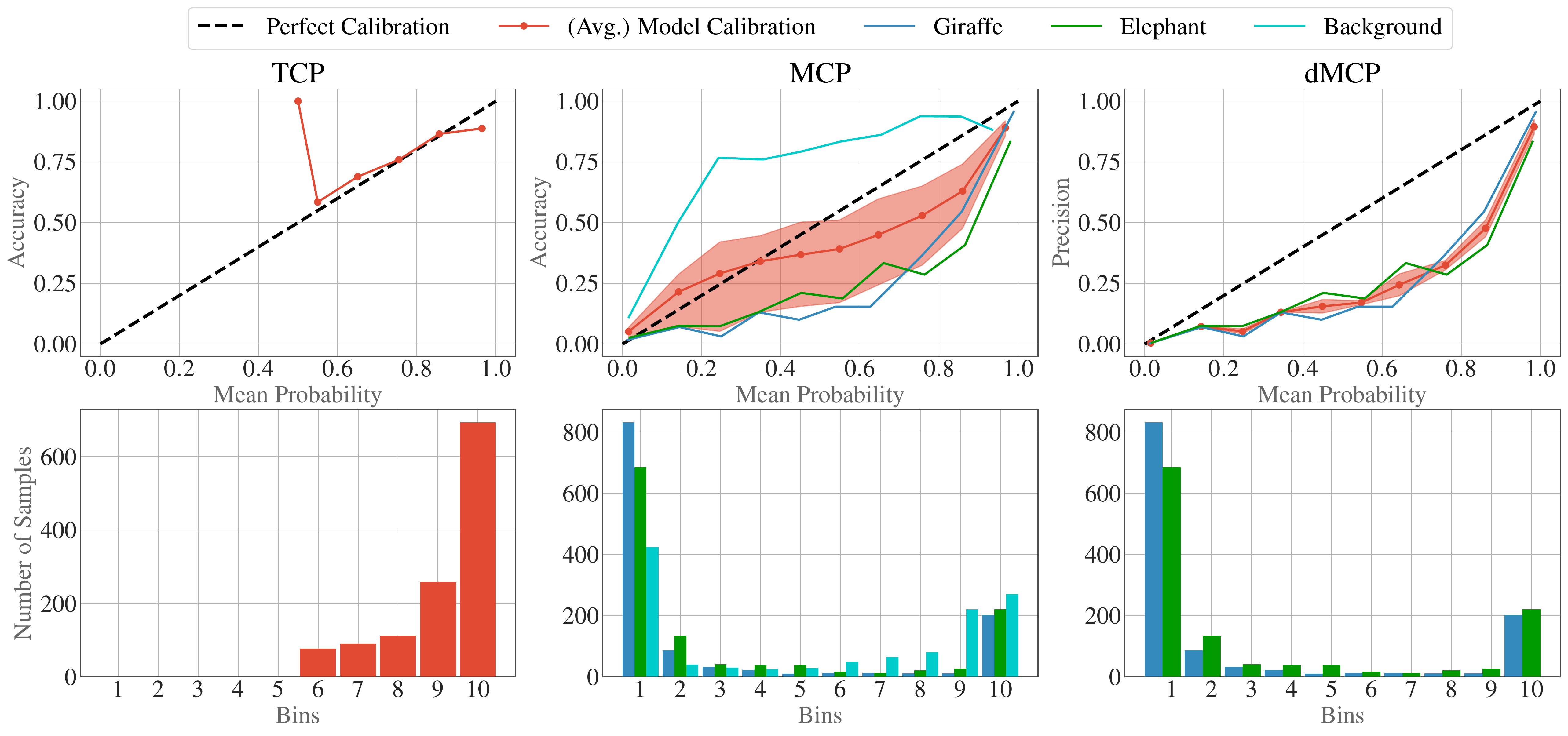}
        \caption{\textsc{Faster-RCNN} on \textsc{animals} subset with post-processing.}
        \label{fig:b_frcnn_animals}
    \end{subfigure}%
    
    \begin{subfigure}{\factor\linewidth}
        \centering
        \includegraphics[width=\linewidth]{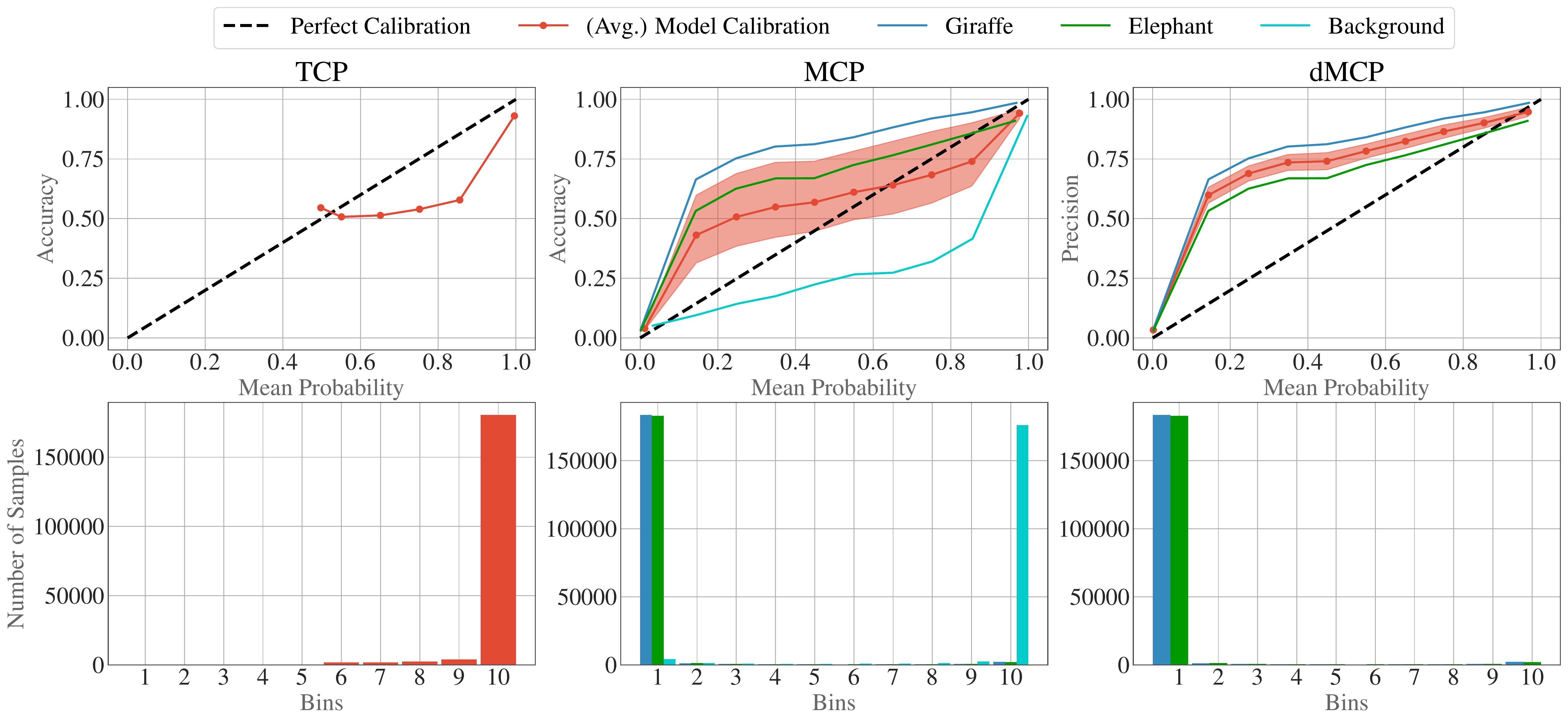}
        \caption{\textsc{Faster-RCNN} on \textsc{animals} subset without post-processing.}
        \label{fig:c_frcnn_animals}
    \end{subfigure}%
    \caption{Calibration plots for the object detection architectures. The red shaded area shows the bin quartiles calculated across classes.}
    \label{fig:calibration_plots}
\end{figure*}
% \begin{itemize}
%     \item How can we build a modular calibration evaluation framework which is suitable for various object detection architectures and ensure an unbiased, fair, and comparable evaluation?
% \end{itemize}

% RQ1: How should we evaluate multi-class object detection architectures regarding their calibration and which metrics should be considered in the evaluation?
%\textit{How should we evaluate multi-class object detection architectures regarding their calibration and which metrics should be considered in the evaluation?}
\rq{Which metrics should be used to evaluate multi-class object detection architectures regarding their calibration and how should these metrics be applied?}
\ra{When evaluating object detection architectures in a multi-class environment, it is essential to assess all predicted probabilities (instead of the maximum one) of our detector by using the MCP and MCE.} 
An example of this is given in the calibration plots shown in Fig.~\ref{fig:calibration_plots} on the \textsc{animals} subset of COCO. The comparison between TCP and MCP demonstrates for all detectors that although their highest predicted probability seems to be fairly well-calibrated, the rest of the predicted probabilities is not. For example, the predictions for the two object classes, giraffe and elephant, are overconfident for \textsc{DETR} and \textsc{Faster-RCNN} with post-processing as their predicted probability lies below the diagonal. Vice versa, when looking at the background class, we can observe underconfidence. Such an interdependency between the object classes and the background class seems reasonable.  In object detection, the neural network needs to learn an extremely imbalanced classification problem between background and objects. Therefore, it makes sense that this potentially biases the quality of our probabilistic predictions either towards under- or overconfidence. Furthermore, as shown by the red curves in Fig.~\ref{fig:calibration_plots}, it is important to not average the bin metrics across classes as this might lead to biased results improving the calibration performance. To further highlight the importance of a suitable metric, we also report the dMCP, which we can see on the right in Fig.~\ref{fig:calibration_plots}. Using precision instead of accuracy, we cannot capture the interdependency between background and objects. This is because the background predictions are ignored by only considering TPs and FPs for evaluation. Accordingly, we observe in Table~\ref{tab:results} that the MCE reports the highest error compared to TCE and dMCE. Note the importance of the number of predictions in a bin when examining the MCP. Comparing only the upper plots of the MCP, \textsc{Faster-RCNN} with post-processing appears to be better calibrated than \textsc{DETR}. However, it also has many predictions in bins between 0 and 1, leading to a higher error, as seen in Table~\ref{tab:results}. In summary, it is essential to consider the MCP and MCE to evaluate the multi-class calibration qualities of detectors as they consider TP, FP, TN, and FN.

% RQ2: How do post-processing steps in a detection pipeline influence the calibration and do architectures that avoid them deliver better calibrated predictions?
\rq{How do post-processing steps in a detection pipeline influence the calibration and do architectures that avoid them deliver better calibrated predictions?}
\ra{The post-processing steps in a detection pipeline significantly influence a model's calibration quality.} We can see an example of this by comparing the calibration plots of \textsc{Faster-RCNN} with (Fig.~\ref{fig:calibration_plots}~(b)) and without (Fig.~\ref{fig:calibration_plots}~(c)) post-processing. When using post-processing steps, the probabilities for object classes (i.e., giraffe and elephant) are overconfident, and the background class probabilities are underconfident. Without post-processing, however, this property no longer holds. There, the probabilities for object classes become underconfident, whereas the probabilities for the background class become overconfident. This means that post-processing steps of \textsc{Faster-RCNN} modify the neural network's probabilistic predictions such that underconfidence of object classes changes to overconfidence. 
Table~\ref{tab:results} also shows that we get much better calibration properties on all datasets from our neural network when avoiding post-processing.  Again, we see that the averages across classes (red curves) would not be sufficient to identify this property.

\ra{Furthermore, we see that architectures avoiding post-processing steps, such as \textsc{DETR}, provide better calibrated probabilistic predictions.} Table~\ref{tab:results} demonstrates for all datasets that the proper scoring rules (NLL and BS) and calibration errors (TCE, MCE, and dMCE) of \textsc{DETR} are below the ones of \textsc{Faster-RCNN} with post-processing. When looking at the calibration plots, this looks surprising as \textsc{Faster-RCNN}'s mean probabilities seem to be closer to the diagonal. However, its predictions are more spread across bins, weighing calibration errors in the middle higher. \textsc{DETR}'s predictions, on the other hand, are concentrated near the edges, which results in overall lower errors. These properties are often referred to in the literature as sharpness and reliability~\citep{ovadia2019can}.

% RQ3: How well are the class probabilities of pretrained object detectors calibrated on samples originating from the training distribution as well as shifted versions thereof?
\rq{How well are the class probabilities of pretrained object detectors calibrated when the detectors are applied on samples from shifted versions of the training distribution?}
\ra{Based on our experiments, the calibration qualities of the pretrained detectors on a dataset based on a shifted version of original the training sample distribution seem to depend on the difficulty of the object detection problem.} Generally, we would expect that the calibration quality worsens as we shift. However, it is also possible that it improves the detector calibration. For example, consider the three-class scenario of the \textsc{animals} subset (COCO vs.~Open Images) in Table~\ref{tab:results}. Here, we observe that both \textsc{DETR} and \textsc{Faster-RCNN} achieve a slightly better mAP, i.e., their performance is better on the shifted test dataset. We believe this quite surprising result is due to the fact that the shifted version of the \textsc{animals} subset contains images on which objects are easier to detect (approx.~50\% bigger objects and fewer objects per image). Similarly, we also see that the calibration errors and proper scoring rules report better results for all detectors. Hence, we conclude that a shift of the sample distribution, which simplifies the problem, can lead to an improved calibration quality of our detections. Looking at the more complex problem \textsc{traffic}, we see that the mAP and the calibration metrics worsen when evaluating them under the shifted version. This is expected and matches the results from literature~\citep{ovadia2019can,harakeh2021estimating}.
The entropy histograms in Fig.~\ref{fig:entropy_hists} show no noticeable difference between the in-distribution dataset of \textsc{traffic} and its shifted version for all architectures. Although the calibration qualities deteriorate, the detectors cannot identify this shift. Optimally, the worse calibration should be reflected by the dissimilarity of these histograms.

% Mean size of boxes in coco: 51898
% Mean size of boxes in open-images: 187380

% \begin{figure*}
%     \centering
%     \includegraphics[width=.9\linewidth]{images/shifted_detr_frcnn.pdf}
%     \caption{Caption}
%     \label{fig:shifted_calibration_plots}
% \end{figure*}

% RQ4: How well is a detector able to identify new objects and out-of-distribution data by means of its probabilistic outputs?
\rq{How well is a detector able to identify new objects in the case of out-of-distribution data by means of its probabilistic outputs?}
\ra{Based on our experiments, pretrained object detectors cannot identify new objects through their probabilistic outputs as they only predict the background class with high probability.} To evaluate this, we visualize log entropy histograms for in-distribution, shifted, and out-of-distribution datasets in Fig.~\ref{fig:entropy_hists}. Instead of the entropy, we use its logarithm to better demonstrate the differences in distributions better. We can see for \textsc{DETR} and \textsc{Faster-RCNN} without post-processing that there are some distributional differences. However, these are due to the fact that the pretrained detectors predict actual object classes  (e.g., giraffe or elephant) for the in-distribution and shifted dataset. In contrast, for the OOD dataset, the detector solely predicts the background class with high probability. As a result, we notice that the predictions on the OOD dataset are even more overconfident when compared to the other histograms. Unfortunately, we cannot leverage this distributional difference to simply detect out-of-distribution samples. 
For instance, consider an autonomous vehicle in front of a flock of sheep. Since our detectors only predict the background class, we can not distinguish this situation from an empty road. Thus, we cannot identify out-of-distribution samples as detectors will just predict the background class.

\section{Simple Recalibration}\label{sec:recal}
\begin{figure*}
    \centering
    \includegraphics[width=\linewidth]{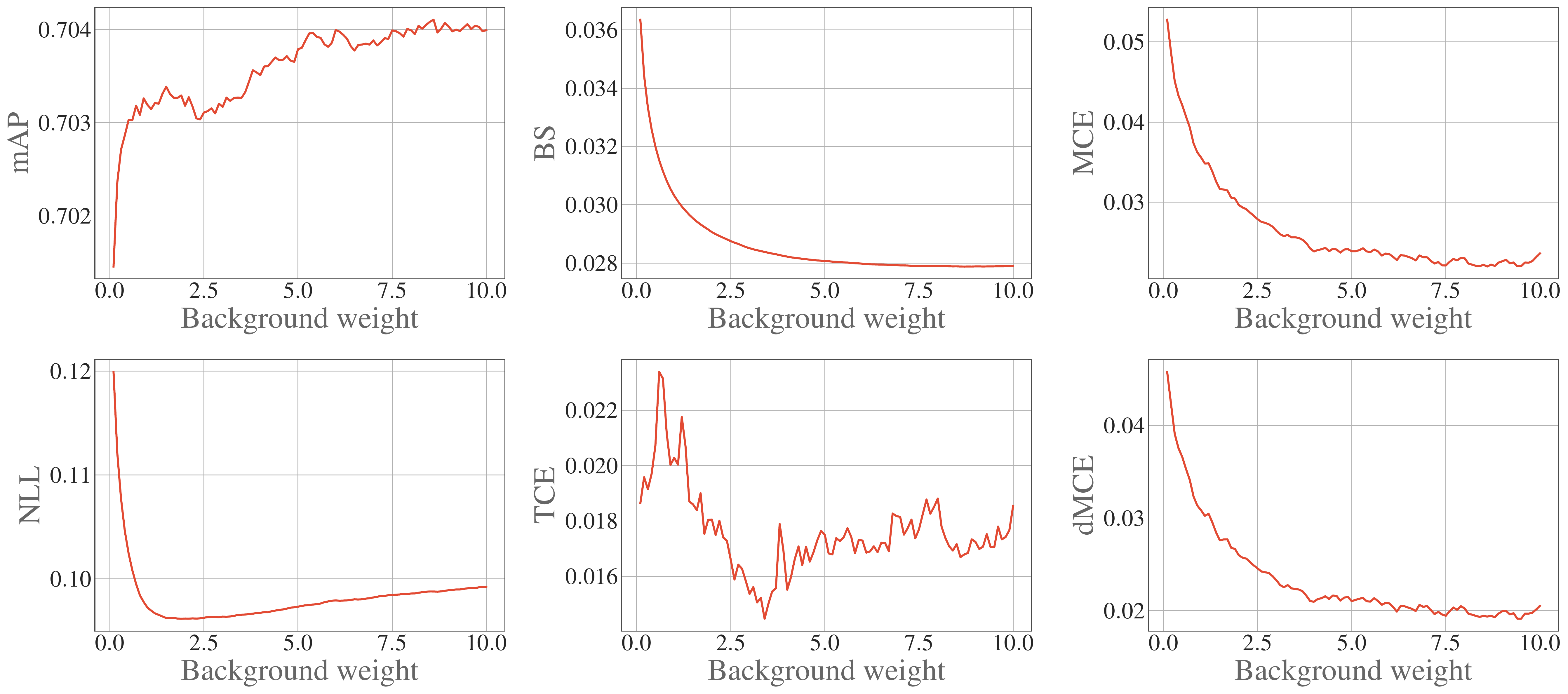}
    \caption{Generalization and calibration metrics with respect to the background weight. Note the different scaling range of the respective metrics. The default background weight (0.1) is the minimum value on the x-axis.}
    \label{fig:results_recal_study}
\end{figure*}
\begin{figure}
    \centering
    \includegraphics[width=\linewidth]{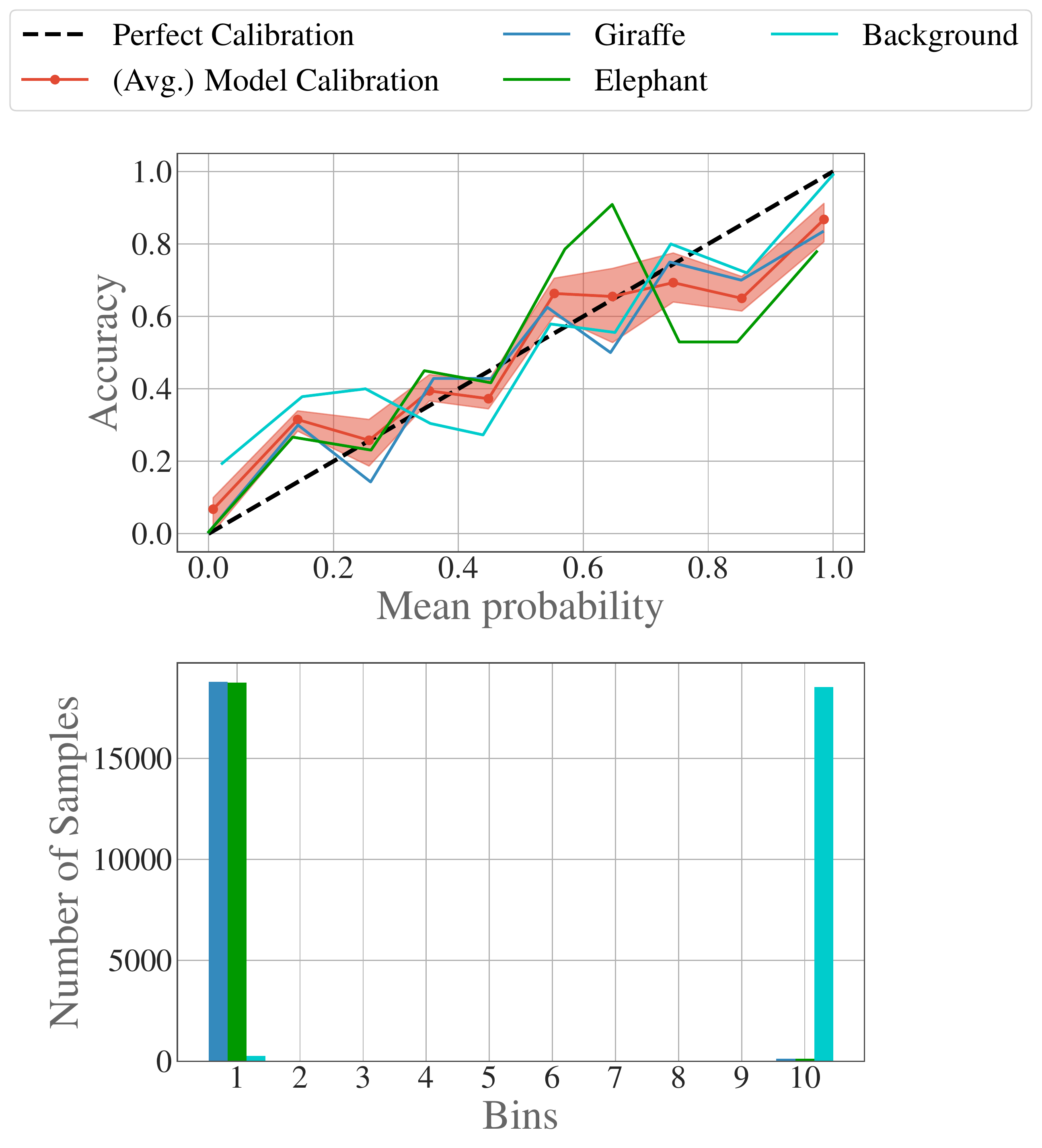}
    \caption{The MCP for the detector that achieved the best MCE (value of $8.6$ for the background weight) on the \textsc{animals} subset.}
    \label{fig:cp_recal_study}
\end{figure}

As mentioned in Sect.~\ref{sec:experiments}, there seems to exist some interdependency between object classes and the background class in a trained detector. While the predictions for the background class are underconfident, the ones for the object classes are overconfident. To investigate this and potentially improve a model's calibration, we conduct a case study on \textsc{DETR} and the \textsc{animals} subset in which we adjust its hyperparameter for the importance of the background class by changing its classification weight during fine-tuning. Note that we can do this for any architecture by simply increasing the weight for the background class in the cross-entropy loss part of a objective function. Specifically, as the background predictions were underconfident, we finetuned 100 object detectors with higher background weights and plot the results in Fig.~\ref{fig:results_recal_study}. We can see that the model's calibration errors and proper scoring rules are improving by increasing the importance of the background class. Furthermore, we also note a slightly increasing generalization performance (mAP). Thus, it seems to be promising for calibration and generalization to raise the background class importance during fine-tuning of a specific detector. In contrast to the other metrics, the TCE does not capture this improvement, highlighting its inappropriateness for the calibration evaluation of multi-class problems again.  Figure~\ref{fig:cp_recal_study} shows the MCP from the detector that achieved the best MCE in this study. Both object and background predictions are closer to the diagonal compared to Fig.~\ref{fig:calibration_plots}~(a) while their interdependencies are no longer recognizable.
We conclude that there seems to be a correlation between calibration quality and the imbalanced problem of object detection (i.e., background vs. objects) and leave further analyses regarding this for future work. 

\section{Conclusion and Outlook}\label{sec:conclusion}
This work assessed different pretrained object detectors regarding their capability of modeling uncertainty considering various factors. First, we proposed a modular framework for evaluating calibration properties of object detection architectures in a multi-class setting while considering all actual and missing predictions of a detector. We analyzed the importance of the choice of metrics and concluded that, besides proper scoring rules, MCP and MCE are the most appropriate metrics for evaluation. Subsequently, we investigated the influence of post-processing steps (e.g., NMS), which worsened the detector's calibration and revealed an interdependency between the confidences of the object class and background class predictions.
We also evaluated the detectors' calibration on datasets with changed distributions. When considering a shift in the sample distribution, we noticed that the calibration does not necessarily worsen as the detection problem may get easier. Furthermore, we saw that detectors could not identify OOD objects when considering an OOD dataset as they only predicted the background class with high probability.
Finally, based on the interdependency insights, we conducted a case study in which we demonstrated that \textsc{DETR}'s calibration can be improved by simply increasing the importance weight of the background class in the objective function during training. 

% Additional detection paradigms
For future work, we need to \textbf{extend the presented framework} such that it includes additional metrics (box-dependent calibration errors~\citep{kuppers2020multivariate}) or we can evaluate additional object detection paradigms. Currently, we assume that the neural network predictions in the classification task describe the parameters of a categorical distribution. This assumption, however, does not always hold. For example, the architecture Retinanet, which is a one-stage paradigm, assumes that every prediction describes a two-class problem between object and background. Hence, the network uses the binary cross-entropy loss function during training. To evaluate such an architecture, we need to extend our framework.

% Avoid post-processing (bad influence on calibration)
As seen in our experiments, post-processing steps such as NMS deteriorate the calibration quality of detectors. Therefore, it is vital to \textbf{research the dependency between post-processing and uncertainty modeling} if we want to use these architectures in tasks such as active learning. For example, it is hard to tell whether it makes sense to enhance an architecture with uncertainty modeling techniques such as Monte-Carlo Dropout~\cite{gal2016dropout} if the outputs are discarded anyway. Therefore, we need to examine whether we can use the direct predictions of a neural network instead of the ones from the detection pipeline or develop more suitable post-processing steps. Additionally, it seems reasonable to just avoid post-processing architectures and use ones such as \textsc{DETR}. They seem promising as we can intuitively recalibrate the predictions of neural networks instead of a detection pipeline that might ignore half of the predictions.
 
% Study interdependency (background and object classes)
Additionally, further research is needed that focuses on the \textbf{class-imbalance problem in object detection} (i.e., interdependency between calibration of object classes and background class) and its \textbf{influence on the uncertainty modeling} of a detector. In our experiments, we solely employed pre-trained detectors and finetuned them on specific subsets. However, it would be interesting to investigate proper scoring rules and calibration errors when training a detector from scratch with different hyperparameter settings.
We assume that training the entire detector with a very high background weight would probably not lead to good generalization.

% Improve OOD detection modeling on potential objects
At last, we want to address the \textbf{missing ability of detectors to identify OOD objects}. We believe that it is vital for many tasks to distinguish between potential unknown objects and a natural background. With that addition, it would be possible, for instance, to improve the responses to different autonomous driving scenarios or perform better exploration during active learning~\citep{herde2021survey}. The training of such an object detection architecture might be realized by using out-of-distribution data~\citep{huseljic2021separation}.

\bibliography{uod}% common bib file


% \begin{figure*}
%     \begin{subfigure}{\linewidth}
%         \centering
%         \includegraphics[width=\linewidth]{images/datasets_shifted/animals_shifted.pdf}
%         \subcaption{Caption}
%     \end{subfigure}
%     \begin{subfigure}{\linewidth}
%         \centering
%         \includegraphics[width=\linewidth]{images/datasets_shifted/traffic_shifted.pdf}
%         \subcaption{Caption}
%     \end{subfigure}
%     \begin{subfigure}{\linewidth}
%         \centering
%         \includegraphics[width=\linewidth]{images/datasets_shifted/ood.pdf}
%         \subcaption{Caption}
%     \end{subfigure}
%     \label{fig:my_label}
% \end{figure*}

\end{document}